\documentclass[10pt,twocolumn,letterpaper]{article}

\usepackage{icb}
\usepackage{times}
\usepackage{epsfig}
\usepackage{graphicx}
\usepackage{amsmath}
\usepackage{amssymb}

\usepackage[hyphens]{url}
\usepackage[caption=false,font=footnotesize]{subfig}
\usepackage[flushleft]{threeparttable}
\usepackage{multirow}
\usepackage{fancyhdr}

\icbfinalcopy % *** Uncomment this line for the final submission

\ificbfinal\fi
\begin{document}

%%%%%%%%% TITLE
%\title{Spoof vs Live Fingerprints: Learning a One-Class Classifier}
\title{Generalizing Fingerprint Spoof Detector: Learning a One-Class Classifier}

\author{Joshua J. Engelsma and Anil K. Jain\\
Michigan State University\\
East Lansing, Michigan, USA\\
{\tt\small \{engelsm7, jain\}@cse.msu.edu}
}

\maketitle

\thispagestyle{empty}

%%%%%%%%% ABSTRACT
\begin{abstract}
   Prevailing fingerprint recognition systems are vulnerable to spoof attacks. To mitigate these attacks, automated spoof detectors are trained to distinguish a set of live or bona fide fingerprints from a set of known spoof fingerprints. Despite their success, spoof detectors remain vulnerable when exposed to attacks from spoofs made with materials not seen during training of the detector. To alleviate this shortcoming, we approach spoof detection as a one-class classification problem. The goal is to train a spoof detector on only the live fingerprints such that once the concept of ``live" has been learned, spoofs of any material can be rejected. We accomplish this through training multiple generative adversarial networks (GANS) on live fingerprint images acquired with the open source, dual-camera, 1900 ppi RaspiReader fingerprint reader. Our experimental results, conducted on 5.5K spoof images (from 12 materials) and 11.8K live images show that the proposed approach improves the cross-material spoof detection performance over state-of-the-art one-class and binary class spoof detectors on 11 of 12 testing materials and 7 of 12 testing materials, respectively.
\end{abstract}

%%%%%%%%% BODY TEXT
\section{Introduction}

\newcommand{\specialcell}[2][c]{%
  \begin{tabular}[#1]{@{}c@{}}#2\end{tabular}}
  \newcommand{\tabitem}{~~\llap{\textbullet}~~}

Automated fingerprint identification systems continue to proliferate into many different domains around the globe, including forensics, border crossing security, national ID systems, and mobile device access and payments~\cite{handbook}. While these systems have become widely accepted due to their accuracy, speed, and purported security, many studies have shown that the systems are highly vulnerable to spoof attacks~\cite{gummy1, gummy2}. Successfully carrying out a spoof attack can be as simple as 2D printing with conductive ink to replicate the fingerprint of a victim left behind on their keyboard.

\begin{figure}[t]
\begin{center}
\includegraphics[scale=0.7]{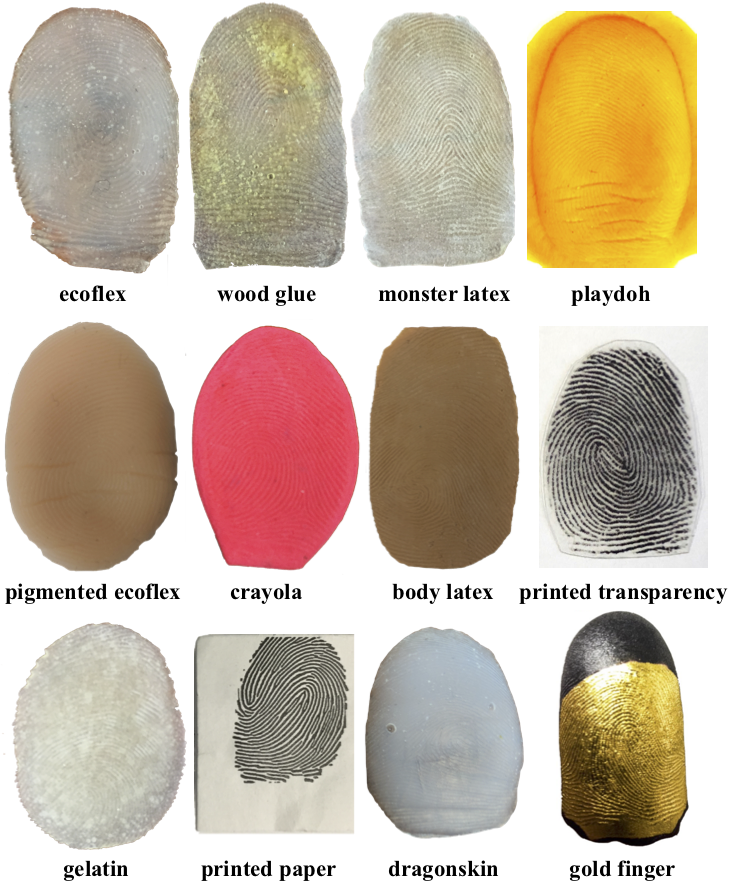}
\caption{Example spoof artifacts. Spoof attack materials have a large variety of both optical and mechanical properties. As such, many spoof detectors which are trained on some subset of spoofs tend to fail to detect spoofs made from materials not seen during training.}
\label{fig:intro_fig}
\end{center}
\vspace{-1.0em}
\end{figure} 

Due to the magnitude of this security flaw existing in many prevailing fingerprint recognition systems, automated spoof detection systems utilizing both hardware and software have been developed over the last several decades with the goal of automatically detecting and flagging spoof attacks\footnote{\url{https://www.youtube.com/watch?v=fZJI_BrMZXU}} prior to performing biometric authentication\footnote{The IARPA ODIN program is an ongoing US government initiative aimed at developing robust spoof detection systems.~\url{https://www.iarpa.gov/index.php/research-programs/odin}}.  Hardware based approaches to spoof detection involve adding additional sensors to the fingerprint reader with the goal of capturing features such as heartbeat, thermal output, blood flow, odor, and sub-dermal fingerprints better able to distinguish live fingerprints from spoof fingerprints~\cite{survey}. In contrast to hardware based approaches, software based solutions do not require additional sensors. Rather, they extract textural, anatomical, physiological, or learned~\cite{CNN3} features from the same image which is to be used for biometric authentication. These features are then used to train a classifier to separate live fingerprints from spoof fingerprints~\cite{survey}. 

Most all spoof detectors, whether hardware based or purely software based, make the assumption that spoof detection is a binary, closed-set problem (i.e. live or spoof). In reality, because spoofs can be fabricated using many different materials with different optical and mechanical properties~(Fig.~~\ref{fig:intro_fig}), spoof detection is an open-set classification problem with an unknown number of ``spoof" classes. Indeed, several studies have shown up to a three-fold increase in error when testing spoof detectors with spoofs made from ``unseen" materials~\cite{inter1, inter2}.

\begin{figure}[t]
\begin{center}
\includegraphics[scale=0.55]{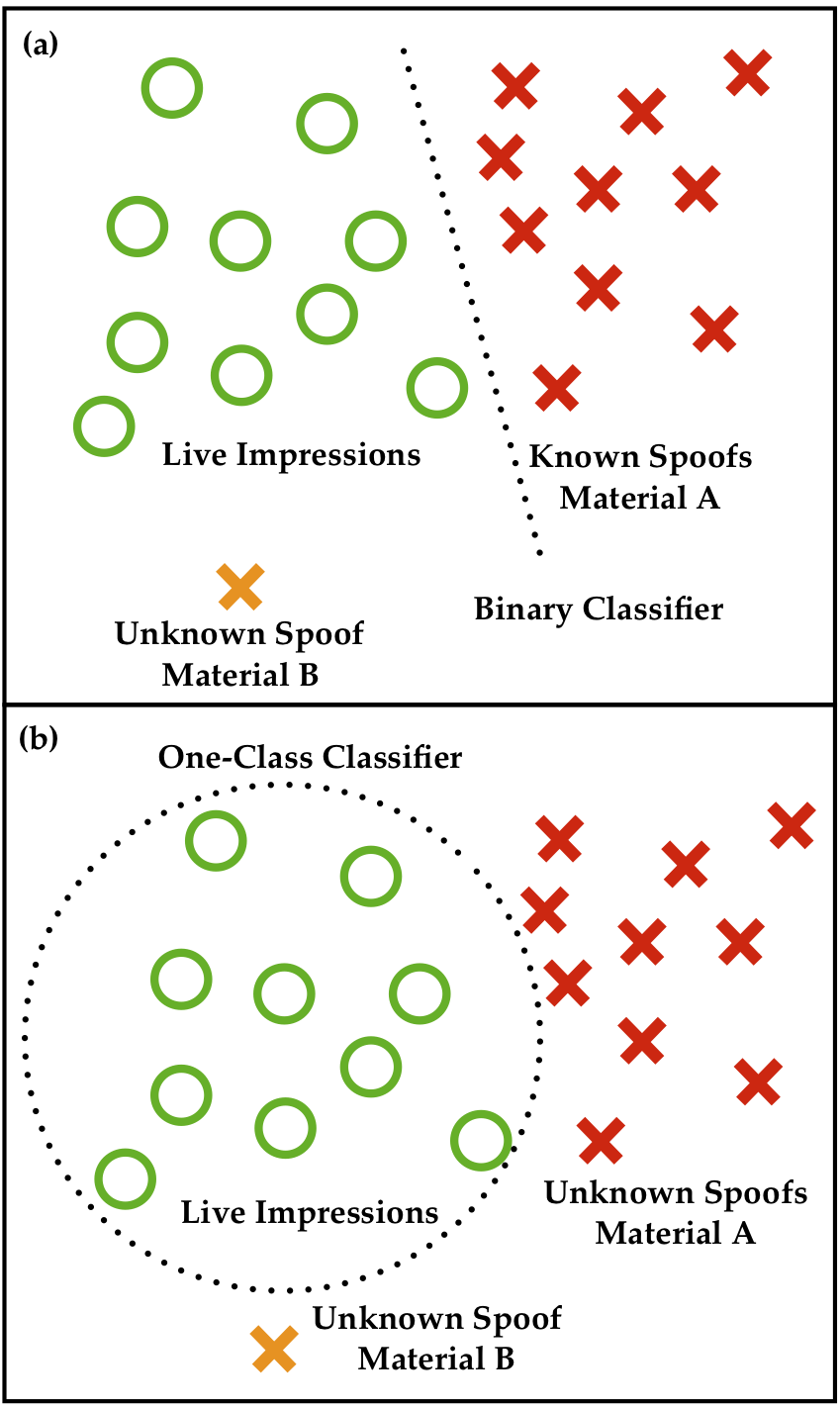}
\caption{Toy example of both two-class (training with both live and spoof) and one-class (training with only live) spoof detectors. (a) A spoof detector trained as a binary classifier is prone to overfitting to ``known materials" seen during training. An ``unknown spoof" made from a material not seen during training may still fall far from the live distribution but be incorrectly classified based on the learned decision boundary. (b) In contrast, the one-class classifier learns a tight decision boundary around the live samples correctly classifies all of the spoofs (none of which have been seen during training).}
\label{fig:overview}
\end{center}
\vspace{-1.5em}
\end{figure}

Some ongoing studies~\cite{open1, open3} have attempted to address the cross-material failures of spoof detectors by explicitly approaching spoof detection as an open-set classification problem using novel material detection~\cite{open1} or one-class classification~\cite{open3}. Although, these studies do not meet the requirements for field deployments with error rates in access of 20\% we are motivated by the tremendous potential of the one-class classification approach in~\cite{open3}. In particular, one-class classification offers the following main advantages over binary classifiers (2-class) in the context of fingerprint spoof detection:

\begin{enumerate}
\itemsep0em
\item Only live samples are needed for training the detector. This eliminates the arduous task of fabricating and imaging a large number of spoof impressions from multiple materials. 
\item One-class classifiers do not overfit to spoof impressions of a particular material during training (as binary classifiers are prone to)~(Fig.~\ref{fig:overview}) such that the cross-material performance decreases. Indeed, one-class classifiers only learn what constitutes a live fingerprint and do not use spoof impressions of \textit{any} specific material or subsets of materials during training. 
\end{enumerate}

\begin{figure}[t]
  \centering
  \subfloat[]{\includegraphics[scale=.35]{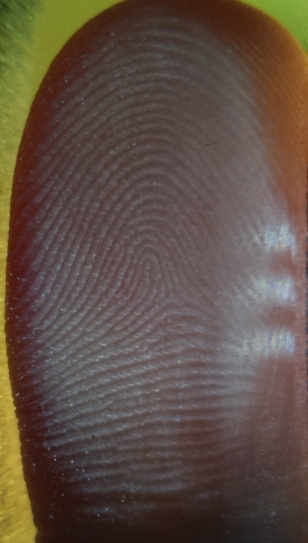}\label{fig:josh_direct}}
  \hfill
  \subfloat[]{\includegraphics[scale=.35]{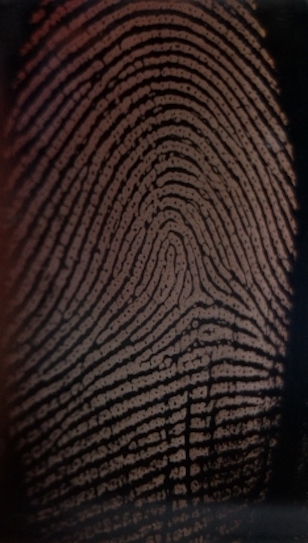}\label{fig:josh_ftir}}
  \caption{Example RaspiReader live impression images. By mounting two cameras at different angles to a glass prism, RaspiReader captures (a) a RGB direct-view image of the finger in contact with the platen; (b) a RGB raw frustrated total internal reflection (FTIR) fingerprint. Both 1900 ppi images contain complementary information useful for spoof detection. Figure retrieved from~\cite{match}.}
  \label{fig:images}
  \vspace{-0.0em}
\end{figure}

Given the above advantages of one-class classification in comparison to binary classification within the context of spoof detection, we identify and rectify several significant limitations of previous approaches in reformulating fingerprint spoof detection into a one-class classification problem. (i) First, Ding and Ross~\cite{open3} were extracting features from \textit{processed FTIR}\footnote{Frustrated Total Internal Reflection (FTIR) is the optical phenomena used by fingerprint readers to capture light reflected back only from fingerprint ridges, enabling capture of a high contrast fingerprint image. The raw FTIR images (Fig.~\ref{fig:josh_ftir}) are processed (RGB to grayscale, contrast enhancement, perspective correction) into processed FTIR images to improve fingerprint matching performance. Direct-view images~(Fig.~\ref{fig:josh_direct}) capture light reflected back from the entire fingerprint area (ridges and valleys).} images to model the distribution of live samples. However, as shown by Engelsma et al. in~\cite{raspireader} features extracted from processed FTIR images can be very similar for both live and spoof fingerprint images in comparison to RGB \textit{raw FTIR}~(Fig.~\ref{fig:josh_ftir}) and \textit{direct-view} images~(Fig.~\ref{fig:josh_direct}). (ii) The features extracted in~\cite{open3} were handcrafted textural features whereas recent research~\cite{CNN2,CNN3} indicates that \textit{learned} deep features are able to better discriminate between live and spoof fingerprints. (iii) Adequately modeling the live fingerprint distribution (necessary for robust one-class classification using only live samples for training) requires a diverse and large number of live samples. However, the number of live impressions in the LiveDet datasets is relatively small with only 2,000 training impressions (from 81 - 400 fingers of 22 -100 subjects depending on the sensor) collected in a single laboratory environment~\cite{2015}.

\begin{figure}[t]
 \begin{center}
\includegraphics[scale=0.55]{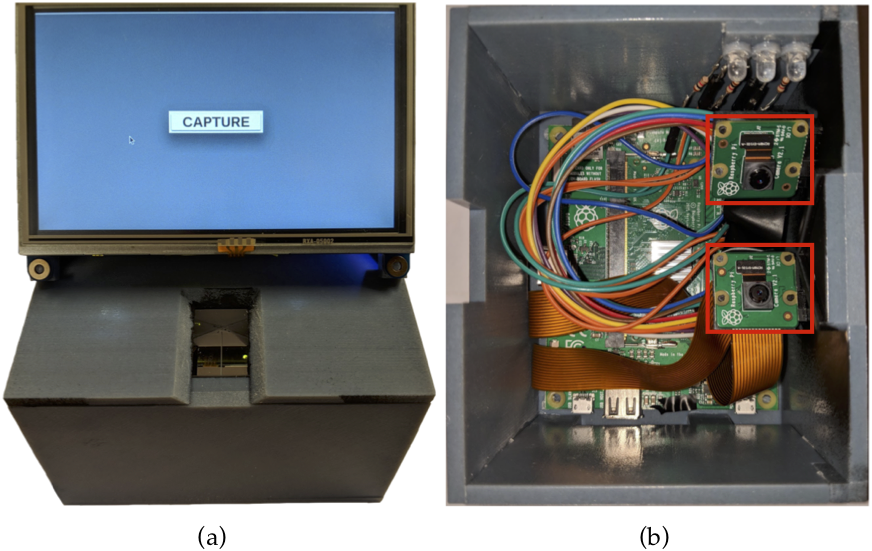}
  \caption{RaspiReader: (a) A prototype of the 4-inch cube RaspiReader (Match in Box model); (b) Internal view of the RaspiReader showing the ubiquitous, low-cost (\$400) components. Both cameras are marked with red boxes. Figure retrieved from~\cite{match}.}
  \label{fig:reader}
  \end{center}
  \vspace{-1.0em}
\end{figure}

\begin{figure*}[t]
\begin{center}
\includegraphics[scale=0.5]{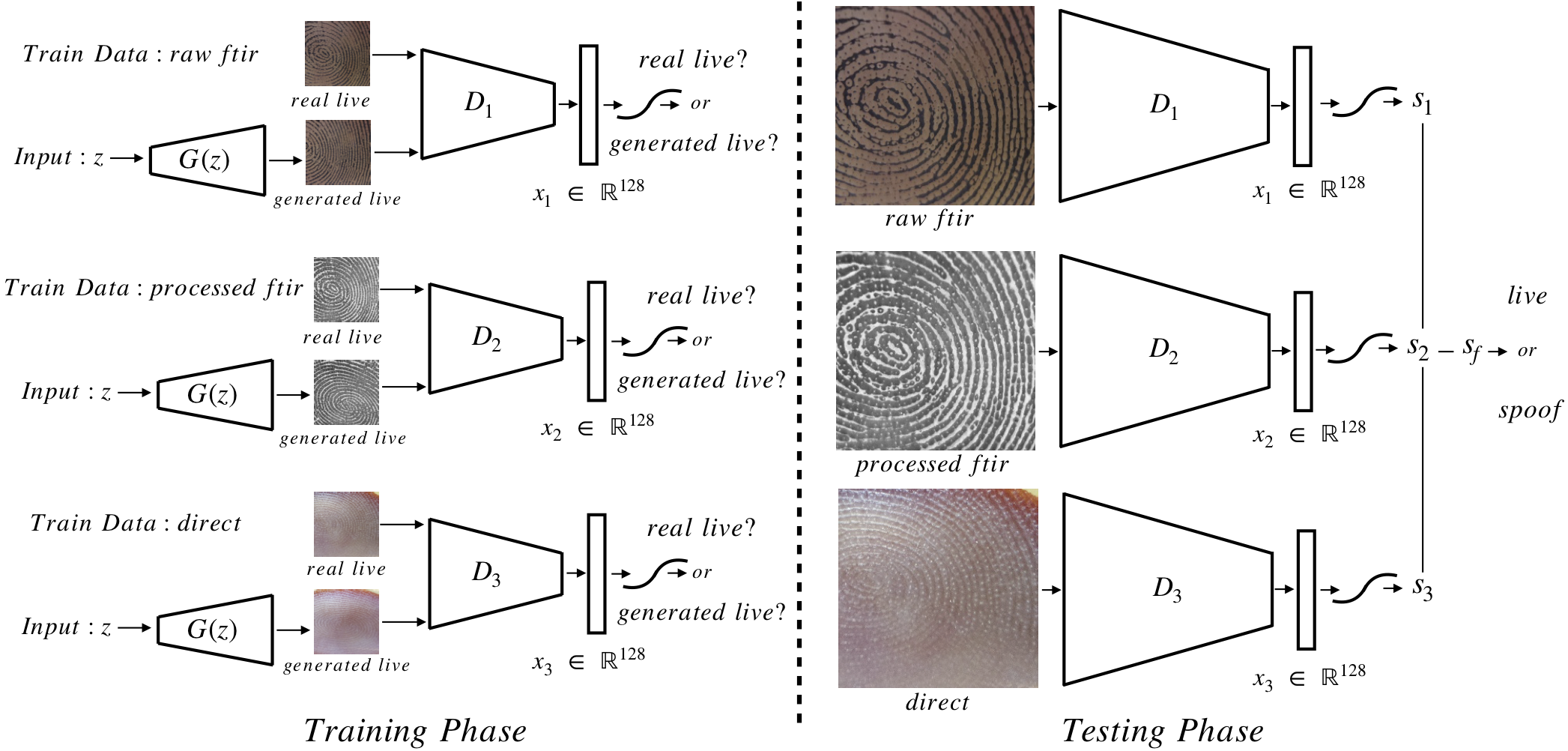}
\caption{Flow diagram of the proposed one-class fingerprint spoof detector. A GAN is trained for each image output of RaspiReader (raw FTIR, processed FTIR, and direct-view). The generator aims to synthesize live fingerprints and the discriminator learns features which separate real live fingerprints from the generator's synthesized live fingerprints. During testing, the features extracted by each discriminator from test images are passed to a sigmoid classification layer to classify the images as live or spoof. The decisions of the three models are fused together by score level fusion.}
\label{fig:schematic}
\vspace{-1.0em}
\end{center}
\end{figure*}

In this paper, we address the aforementioned limitations in~\cite{open3} to improve cross material spoof detection performance. First, we use a RaspiReader fingerprint reader (Fig.~\ref{fig:reader})~\cite{raspireader, match} which uses two cameras to simultaneously capture both a direct-view image and a raw FTIR fingerprint image at 1900 ppi (Fig.~\ref{fig:images}). Using the RaspiReader, we collect a large, diverse dataset comprised of 11,880 live fingerprint impressions from nearly 6,000 unique fingers and 5,531 spoof finger impressions from 12 different materials (Fig.~\ref{fig:intro_fig}). It should be noted that this dataset is significantly larger and more diverse in terms of number of subjects, fingers, materials, and collection locations than any reported LiveDet datasets (585 subjects vs. 100 subjects, 5,800 fingers vs. 400 fingers, 12 materials vs. 6 materials, and 3 locations vs. 1 location). Although we tie our spoof detector specifically to RaspiReader, we note that most existing state-of-the-art spoof detection algorithms are also ``reader specific" since they require re-training for each individual fingerprint reader given the different textural characteristics from one reader to another\footnote{The IARPA ODIN program has specified the Crossmatch Guardian and the Lumidigm as baseline fingerprint readers for software solutions.}.

Next, we train three Generative Adversarial Networks (GANs) on each of the complementary images~\cite{raspireader} output by the RaspiReader (raw FTIR image, processed FTIR image, and direct-view image) (Fig.~\ref{fig:schematic}). For each GAN, (i) The generator attempts to synthesize live fingerprint images. (ii) The discriminator is fed the generator's synthesized live fingerprints and also real live fingerprint images. (iii) The generator learns to synthesize better live fingerprint images as it attempts to fool the discriminator, and the discriminator learns to distinguish between real live fingerprints and synthesized live fingerprints. Our hypothesis is that the features learned by the discriminator to separate real live fingerprints from synthesized live fingerprints can also be used during testing to distinguish live fingerprints from spoof fingerprints. 

After training all three GANS, the generator is discarded and the sigmoid output (probability of an input image being a real live sample) of each discriminator acts as a ``spoofness" score. The fusion of the scores output by all three of the discriminators constitutes the final spoofness score of an input fingerprint sample.

More concisely, the contributions of this work are:

 \begin{table*}[t]
 \small
 \centering
\begin{threeparttable}
\caption{Summary of Spoof Dataset (\# Impressions per Material)}
\label{table:spoof_data}
\begin{tabular}{ |c|c|c|c|c|c|c|c|c|c|c|c|c|c|c|}
 \hline
 Dragonskin & \specialcell{Ecoflex} & \specialcell{Gelatin} & \specialcell{Monster \\Latex}& \specialcell{Crayola \\Magic}& \specialcell{Pigmented \\Ecoflex }& \specialcell{Playdoh}& \specialcell{Woodglue}& \specialcell{Body \\Latex}& \specialcell{2D \\Paper}& \specialcell{Trans-\\parency}& \specialcell{Gold \\Finger}\\
 \hline
 \hline
 96 & 605 & 161& 741& 360& 1040& 348& 501& 861& 556& 50& 212\\
 \hline
\end{tabular}
\end{threeparttable}
\vspace{-1.25em}
\end{table*}

\begin{table}[t]
 \centering
\begin{threeparttable}
\caption{Summary of Live Dataset}
\label{table:live_data}
\begin{tabular}{ |c||c|c|}
 \hline
 Collection Location & \specialcell{\# Subjects\tnote{1}} & \# Impressions\tnote{2} \\
 \hline
 \hline
 Michigan State University & 55 & 3,050\\
 \hline
 Clarkson University & 122 & 1,219\\
 \hline
 Johns Hopkins APL & 408 & 7,611 \\
  \hline
  \hline
  Total & 585 & 11,880\\
  \hline
\end{tabular}
\begin{tablenotes}
\item[1] 10 fingers per subject
\item[2] 1-5 impressions per finger
\end{tablenotes}
\end{threeparttable}
\end{table}

\begin{enumerate}
\itemsep0em
\item The collection (using RaspiReader) of a challenging dataset of 12 different spoof materials and over 11.8K live fingerprints from nearly 6,000 unique fingers.
\item A new ``open-set" spoof detector (i.e. we make no assumptions about the characteristics of spoofs during training). Instead we learn the concept of a live fingerprint such that spoofs of all materials can be rejected. This algorithm is realized using the discriminators of three GANs trained on only live fingerprint images.
\item Experimental evaluation demonstrating the superior performance of our algorithm on this dataset compared to baseline~\cite{open3, raspireader} algorithms.
\end{enumerate}

\section{Approach}

In this section, we (i) provide more details on our dataset, (ii) explain how images are preprocessed via region of interest (ROI) extraction prior to training three GAN networks, (iii) provide details on the architecture and training procedure for each of the GAN networks, and finally, (iv) discuss our fusion technique.

\begin{figure*}[t]
\begin{center}
\includegraphics[scale=0.6]{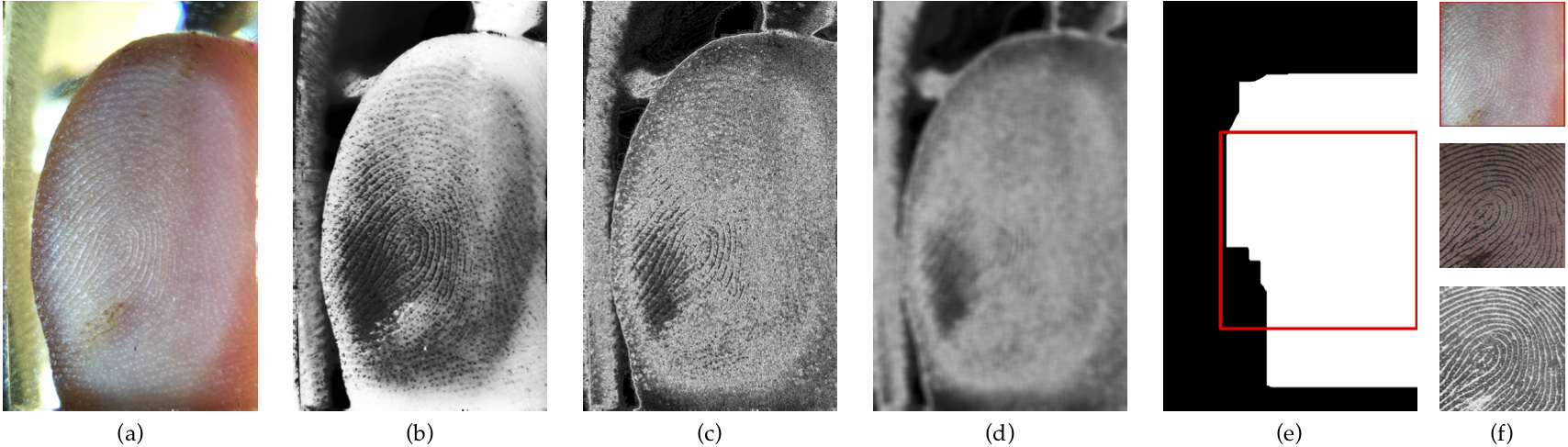}
\caption{Preprocessing. A raw, direct-view fingerprint (a) is converted to grayscale and contrast enhanced (b). Next, the gradient image (c) of (b) is computed using the Laplacian filter and smoothed into (d) using a gaussian filter. Finally, (d) is binarized into (e) and used to locate centroids for cropping ROI patches from the 3 image outputs (direct-view, raw FTIR, and processed FTIR) of RaspiReader (f).}
\label{fig:preprocess}
\vspace{-1.0em}
\end{center}
\end{figure*}

 \begin{table*}[t]
 \small
 \centering
\begin{threeparttable}
\caption{Training, Validation, and Testing Dataset Splits}
\label{table:splits}
\begin{tabular}{ |c|c|c|c|c|c|}
 \hline
 \multirow{2}{*}{\textbf{Live Impressions}} & \textbf{\specialcell{Training and Validation \\ Location\tnote{1}}} & \textbf{\specialcell{Testing \\ Location}} & \textbf{\specialcell{\#Training \\Impressions}}& \textbf{\specialcell{\# Validation\\ Impressions}}& \textbf{\specialcell{\# Testing \\Impressions}}\\
 \cline{2-6}
 & CU \& JHUAPL & MSU &  8,330 & 500 & 3,050\\
 \hline
 \hline
 
\multirow{2}{*}{\textbf{\specialcell{Spoof Partition\tnote{2} \\ $Set_1$}}} & \textbf{\specialcell{Training and Validation \\ Materials}} & \textbf{\specialcell{Testing Materials}} & \textbf{\specialcell{\#Training \\Impressions}}& \textbf{\specialcell{\# Validation\\ Impressions}}& \textbf{\specialcell{\# Testing \\Impressions}}\\
 \cline{2-6}
 & \specialcell{Dragonskin, Ecoflex, \\ Crayola Magic, 2D Paper, \\Body Latex, Monster Latex} &  \specialcell{Gelatin, Playdoh, \\Woodglue, Pigmented Ecoflex, \\Gold Finger, Transparency}  &  2,851 & 134 & 2,312\\
  \hline
  \hline
  
 \multirow{2}{*}{\textbf{\specialcell{Spoof Partition \\ $Set_2$}}} & \textbf{\specialcell{Training and Validation \\ Materials}} & \textbf{\specialcell{Testing Materials}} & \textbf{\specialcell{\#Training \\Impressions}}& \textbf{\specialcell{\# Validation\\ Impressions}}& \textbf{\specialcell{\# Testing \\Impressions}}\\
 \cline{2-6}
 &  \specialcell{Gelatin, Playdoh, \\Woodglue, Pigmented Ecoflex, \\Gold Finger, Transparency} & \specialcell{Dragonskin, Ecoflex, \\ Crayola Magic, 2D Paper, \\Body Latex, Monster Latex}  &  2,195 & 117 & 2,985\\
  \hline
  \hline
  
\end{tabular}
\begin{tablenotes}
\item[1] The location refers to the site at which the fingerprints were collected.
\item[2] The 12 spoof materials were randomly split into two partitions. Only the binary-class baseline utilizes the spoof training data.
\end{tablenotes}
\end{threeparttable}
\end{table*}

\subsection{Dataset}

The dataset used in our experiments consists of 5,531 spoof images from 12 different spoof materials (Table~\ref{table:spoof_data}). Additionally, we have collected 11,880 live fingerprint impressions from over 500 unique subjects (Table~\ref{table:live_data}). In order to most closely mimic a ``real world" scenario, the live fingerprint impressions were captured in three different locations over the course of 6 months, from a diverse population (nearly 6,000 unique fingers compared to 400 in LiveDet).

The datasets in Tables~\ref{table:spoof_data} and~\ref{table:live_data} are further partitioned into training, validation, and testing sets. Because our primary objective is to evaluate cross-material performance (i.e. test on spoofs fabricated from unseen spoof materials), we randomly split our spoof materials into two partitions of 6 materials (unlike our algorithm, the binary-class baseline necessitates spoof training data). Then we evaluate our algorithm and all baselines on each partition ($Set_1$ and $Set_2$) separately~(Table~\ref{table:splits}).

\subsection{Preprocessing}

Prior to training our proposed algorithm, we preprocess the fingerprint images via a region of interest (ROI) extraction algorithm in order to simplify the learning task of the GAN networks. The region of interest extraction also removes noisy background details from the fingerprint images. Removing the noisy background information is important, since live fingerprints and spoof fingerprints will have similar noise patterns in the background (especially in the direct-view images) causing live and spoof images to appear more similar. We validated this claim experimentally, observing that the features were very similar for live and spoof testing samples without preprocessing.

Given a set of three fingerprint images $\{I_{direct}, I_{raw}, I_{processed}\}$ from a single acquisition on RaspiReader, we first find the ROI of the direct-view image ($I_{direct}$). Then, a mapping is found between the ROI of the direct-view image and the ROIs of the raw FTIR image ($I_{raw}$) and processed FTIR image ($I_{processed}$). 

To find the ROI of $I_{direct}$, we perform a number of conventional image processing operations. First, $I_{direct}$~(Fig.~\ref{fig:preprocess}a) is converted to grayscale and contrast enhanced via histogram equalization to $I_{direct, contrast}$~(Fig.~\ref{fig:preprocess}b). Next, the gradient image of $I_{direct, contrast}$ is taken to highlight the fingerprint region of interest from the background~(Fig.~\ref{fig:preprocess}c). Let this gradient image be denoted as $I_{direct, gradient}$. The gradient image is smoothed with a $30~\times~30$ gaussian filter~(Fig.~\ref{fig:preprocess}d), binarized with an intensity threshold of 75, and de-noised with the morphological operations of erosion and dilation into $I_{binary}$. Using $I_{binary}$, we locate the centroid $(c_{direct,x}, c_{direct,y})$ of the foreground ``blob" with the largest area. This centroid is then used as a center point for our alignment window of size $768~\times~768$~(Fig.~\ref{fig:preprocess}e). Finally, a mapping is found between the direct-view ROI computed in~(Fig.~\ref{fig:preprocess}e) and the ROIs of the raw FTIR and processed FTIR images. The ROIs of all three images $\{ROI_{direct}, ROI_{raw}, ROI_{processed}\}$ after extraction and resizing to $256~\times~256$ are shown in Figure~\ref{fig:preprocess}f.

\subsection{GAN Training} 

After extracting $\{ROI_{direct}, ROI_{raw}, ROI_{processed}\}$ from each image in our dataset, we train three GAN networks, one for each of the RaspiReader image types (Fig.~\ref{fig:schematic}). We select the DCGAN architecture proposed in~\cite{dcgan} because unlike other GANs, DCGAN includes a classification loss for training the discriminator. This a natural choice for us, since our end goal is to use the trained discriminator to classify between live fingerprints and spoof fingerprints. 

It is of utmost importance to note that each of the three DCGANs ($DC_{1}$, $DC_{2}$, and $DC_{3}$) used in our algorithm are trained \textbf{only} on live fingerprint images. The learning objective of each GAN is as follows. (i) Each $DC_i,~i \in \{1, 2, 3\}$ is comprised of a generator $G_i$ and a discriminator $D_i$. (ii) The generator $G_i$ takes as input a random vector $z \in \mathbb{R}^{100}$ drawn from a standard multivariate distribution and outputs a synthesized live fingerprint image. (iii) The discriminator $D_i$ takes as input both real live fingerprint images and also synthesized live fingerprint images output by the generator. The discriminator then learns features which separate real live fingerprint images from synthesized live fingerprint images. In this manner, the discriminator learns the features that define real live fingerprint images; this is the ultimate goal of our one-class spoof detection formulation. 

More formally, the DCGAN is optimized in accordance to the adversarial loss function in Equation~\ref{eq:gan} below.
\vspace{-.5em}
\begin{equation}
\label{eq:gan}
\mathcal{L}_{adv}(G, D) =  \mathbb{E}_x~[logD(x)] + \mathbb{E}_z~[log(1-D(G(z)))]
\end{equation} where $G$ is the generator model, $D$ is the discriminator model, $x$ is a sample from the real live distribution, and $z \in \mathbb{R}^{100}$ is a vector drawn from a standard multivariate distribution.

In our experiments, we modified DCGAN to take input fingerprints of size $256~\times~256~\times~3$ for the raw FTIR images and the direct-view images and $256~\times~256~\times~1$ for the processed FTIR images (rather than the default\footnote{\url{https://github.com/carpedm20/DCGAN-tensorflow}} input size of $64~\times~64$). This was accomplished by adding several convolution layers. We further modified the discriminator to have a fully connected layer, outputting compact feature representations of length $128$. In total, our modified discriminator architecture consists of 5 convolution layers (each having $5~\times~5$ filters and a stride of 2), an average pooling layer, and two fully connected layers (128-dimensional for feature representation, followed by 1-dimensional for sigmoid classification layer). Every convolution layer is followed by Leaky Relu activation. Additionally, group normalization is performed after every convolution layer except the first. We found that batch normalization resulted in very unstable spoof detection performance, which was significantly stabilized using group normalization. We note that we also experimented with deeper state-of-the-art architectures for the discriminator such as MobileNet~\cite{mobilenet}, however, the generator did not converge with deeper discriminator models. We train our GANS with a batch size of 64, a learning rate of $0.0002$, and the Adam optimizer.

Note that while we do not use any spoof data to train the GAN network, we do use a small subset of spoofs as a validation set (Table~\ref{table:splits}) to determine when to stop the GAN network training. In this manner, we do make use of the available spoof impressions, rather than neglecting valuable data. At the same time, since we do not use the spoof data for training (only for validation) we avoid the risks of overfitting to spoofs made from a specific subset of materials (as binary classifiers are prone to). Furthermore, the validation set is comprised of a very small number of spoofs ($\approx$~120 impressions), inline with a secondary goal of our approach, namely to develop a spoof detector which does not necessitate the laborious task of creating and imaging an inordinate number of spoofs.

Finally, we also experimented with (i) minutiae patch based GANS (i.e. the input to the GAN network was minutiae patches extracted from the region of interest in each image), and (ii) Variational Auto-Encoder anomaly detectors trained on the region of interests from each image (i.e. the reconstruction error of the VAE was used as an anomaly score). Both of these models are fused into our final score.

 \begin{table*}[t]
 \footnotesize
 \small
 \centering
\begin{threeparttable}
\caption{True Detection Rates (TDR) for Individual Testing Materials from Spoof Partition $Set_1$ and $Set_2$ (FDR = 0.2\%)}
\label{table:s1}
\begin{tabular}{ |c||c|c|c|c|c|c|c|c|c|}
 \hline
 \multirow{2}{*}{\textbf{$Set_1$}} & \textbf{Algorithm} & \specialcell{\textbf{Gelatin}} & \specialcell{\textbf{Pigmented} }& \specialcell{\textbf{Playdoh}}& \specialcell{\textbf{Woodglue}} & \specialcell{\textbf{Transparency}}& \specialcell{\textbf{Gold Finger}}\\
 \cline{2-8}
 \cline{2-8}
 &Texture + OCSVM~\cite{open3} & 0.0\% & 2.0\% & 0.3\% & 0.0\% & 0.0\% & 0.9\% \\
 \cline{2-8}
 &Binary CNN~\cite{raspireader} & 67.7\% & \textbf{29.4\%} & 6.0\% & 55.7\% & 34.0\% & 11.8\% \\
  \cline{2-8}
 &Proposed 1-class GANs & \textbf{74.5\%} & 22.3\% & \textbf{96.3\%} & \textbf{85.2\%} & \textbf{94.0\%} & \textbf{39.2\%} \\
\hline
\hline
 \multirow{2}{*}{\textbf{$Set_2$}} & \textbf{Algorithm} & \textbf{Dragonskin} & \specialcell{\textbf{Ecoflex}} & \specialcell{\textbf{Monster Latex}}& \specialcell{\textbf{Crayola Magic}} & \specialcell{\textbf{Body Latex}}& \specialcell{\textbf{2D Paper}}\\
 \cline{2-8}
 \cline{2-8}
 & Texture + OCSVM~\cite{open3} & 0.0\% & 0.2\% & 0.3\% & 15.0\% & \textbf{20.6\%} & 33.8\% \\
 \cline{2-8}
 & Binary CNN~\cite{raspireader} & \textbf{49.0\%} & \textbf{39.3\%} & \textbf{54.3\%} & 78.1\% & 12.1\% & 46.1\% \\
 \cline{2-8}
 & Proposed 1-class GANs & 2.1\% & 4.8\% & 38.5\% & \textbf{83.6\%} & 0.3\% & \textbf{56.8\%}\\
 \hline
\end{tabular}
\end{threeparttable}
\vspace{-.5em}
\end{table*}

\begin{table}[h]
\footnotesize
\caption{Speed Comparison of Spoof Detectors}
 \centering
\begin{threeparttable}
\begin{tabular}{|c |c |c| c|}
 \hline
 Algorithm & \specialcell{OCSVM~\cite{open3}} & CNN~\cite{raspireader} & Proposed\\
 \hline
 Time (ms) & 1,914 & 472 & 778 \\
 \hline
\end{tabular}
\begin{tablenotes}
\item[1] Computed on 2.9 GHz i5 with 8 GB of RAM
\end{tablenotes}
\end{threeparttable}
\label{table:speed}
\end{table}

\subsection{Score Fusion}

During testing, the generator of each GAN is discarded and the discriminators are used for spoof detection. In particular, given three images $\{I_{direct}, I_{raw}, I_{processed}\}$ from a single RaspiReader acquisition, three scores $\{s_{direct}, s_{raw}, s_{processed}\}$ are obtained from $DC_{1}$, $DC_{2}$, and $DC_{3}$, respectively (Fig.~\ref{fig:schematic}). We further obtain a score $s_{patches}$ from the discriminator of a DCGAN trained on the raw FTIR minutiae-based patches of RaspiReader, and $s_{vae}$ an anomaly score obtained from a VAE trained on the direct view images of RaspiReader. Then, a final spoof detection score $s_f$ is computed as the average of the five scores.

We also experimented with training different one-class classifiers such as One-Class SVMs~\cite{open3, open4} and Gaussian Mixture Models~\cite{open5} on top of the 128-dimensional features extracted by the discriminators given their prior use in fingerprint and face spoof detection, but found no performance improvement over directly using the sigmoid output of the discriminators.

\section{Experimental Results}

Our experimental results aim to demonstrate the superior cross material performance of our proposed one-class spoof detector in comparison to state-of-the-art one-class spoof detectors and binary-class spoof detectors. Note that, our proposed algorithm and the baseline one-class spoof detector proposed in~\cite{open3} do not use spoof data for training, but do use the validation spoof data listed in Table~\ref{table:splits} to determine when to stop training and / or tune hyper-parameters during a \textbf{validation phase}. In contrast, the binary-class baseline algorithm in~\cite{raspireader} requires spoof data for training. 

We also note that we slightly modified~\cite{open3} in order to boost its performance to make for a fair comparison with our proposed algorithm. In particular, the algorithm in~\cite{open3} extracted 5 different textural features from grayscale processed FTIR images to train 5 one-class support vector machines (OCSVMs) for one-class spoof detection. Since RaspiReader outputs 3 images, we extract the same 5 textural features used in~\cite{open3} but from all 3 images. In total then we trained 15 OCSVMs (3 images and 5 OCSVMs per image) and fused their scores. Also, since RaspiReader outputs RGB images, rather than extracting grayscale textural features as was done in~\cite{open3}, we extract color-textural features (i.e. the texture feature is extracted from all 3 channels and concatenated into a final feature vector) from the raw FTIR and direct-view images. Making these modifications increased the computational time of the algorithm (Table~\ref{table:speed}), but boosted the spoof detection performance in all scenarios.

\subsection{Analysis}

For 11 of the 12 spoof materials (Table~\ref{table:s1}), our proposed algorithm outperforms the one-class baseline algorithm from~\cite{open3} and for 7 of the 12 spoof materials our algorithm outperforms the binary-class baseline algorithm from~\cite{raspireader}. The average spoof detection performance of our proposed algorithm across all twelve of the materials is $TDR = 49.8\%$ compared to that of the best baseline algorithm (binary-CNN) at $TDR = 40.3\%$.

\begin{figure}[t]
\vspace{-1.0em}
  \centering
  \subfloat[]{\includegraphics[scale=.43]{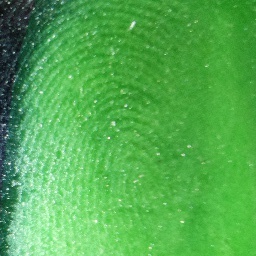}\label{fig:playdoh_direct}}
  \hfill
  \subfloat[]{\includegraphics[scale=.43]{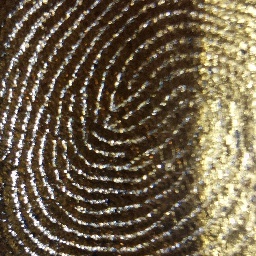}\label{fig:gold_direct}}
  \caption{Green playdoh finger direct-image (a); gold finger direct-image (b). Even though qualitatively these images look very different from live fingers, the binary classifier which has not seen playdoh or gold finger during training misclassifies them as live fingers. In contrast, our proposed GANs are able to easily reject these very anomalous spoofs.}
  \label{fig:failures_binary}
  \vspace{-0.5em}
\end{figure}

\begin{figure}[t]
  \centering
  \subfloat[]{\includegraphics[scale=.43]{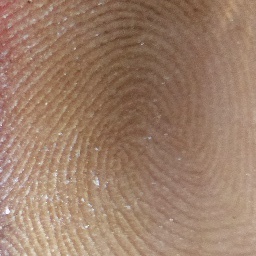}\label{fig:live_direct}}
  \hfill
  \subfloat[]{\includegraphics[scale=.43]{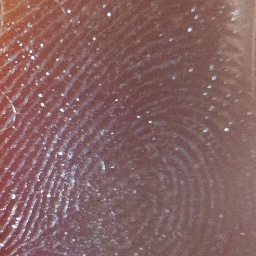}\label{fig:ecoflex_direct}}
  \caption{Live finger direct-image (a); ecoflex finger direct-image (b). Both baselines as well as our GANs struggle to distinguish clear spoofs like ecoflex from live fingers, since much of the live finger can be seen from behind the spoof.}
  \label{fig:failures_gan}
  \vspace{-0.0em}
\end{figure}

\begin{figure}[t]
\begin{center}
\includegraphics[scale=0.6]{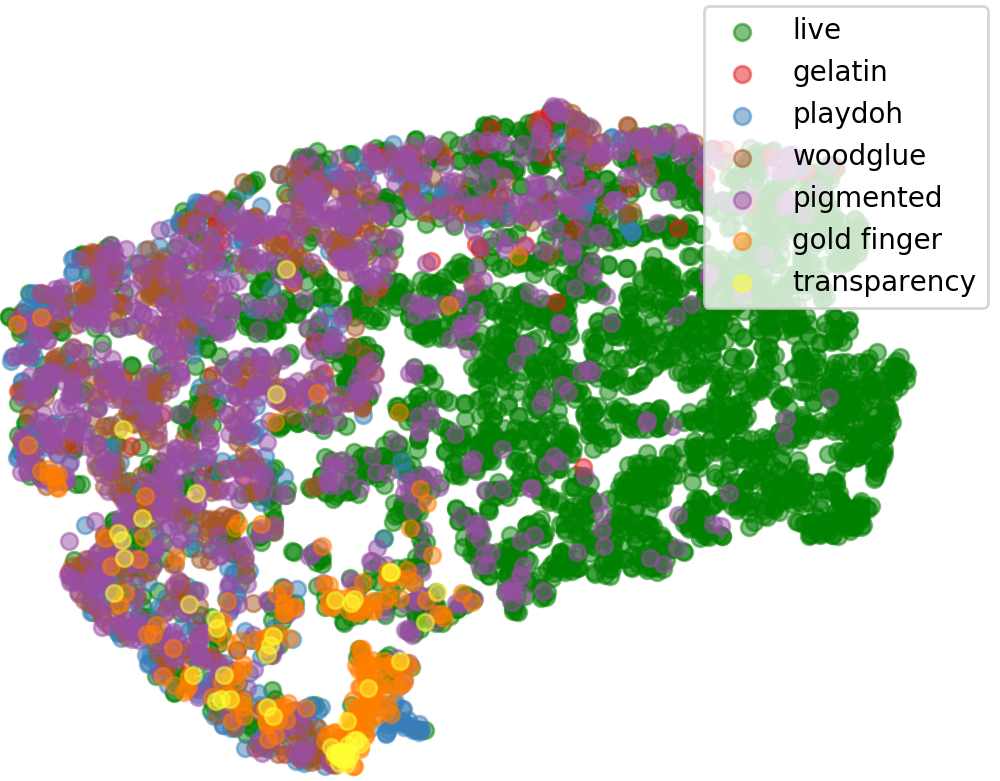}
\caption{2D t-SNE visualization of the 384-dimensional (128 features for each model) live and spoof representations extracted by our three GANS from the live testing data and spoof testing data from partition $Set_1$. The most difficult material from $Set_1$, pigmented ecoflex, shares some overlap with the live distribution.}
\label{fig:tsne}
\vspace{-1.0em}
\end{center}
\end{figure}

The baseline algorithm proposed in~\cite{open3} performs very poorly in all testing scenarios. While the classification accuracy for some materials using~\cite{open3} was only slightly lower than the classification accuracy reported in~\cite{open3}, the $TDR$ at the fixed operating of $FDR = 0.2\%$ was significantly lower\footnote{FDR = 0.2\% is the IARPA ODIN specified threshold.}. This demonstrates that the handcrafted textural features used by the authors in~\cite{open3} are not discriminative enough for training one-class classifiers for state-of-the-art spoof detection. In contrast, the deep features extracted by our proposed approach are discriminative enough to separate live and spoof fingerprints, even when training with only live fingerprints (Fig.~\ref{fig:tsne}).

Next, we note that the binary-CNN baseline performance is lower than the cross material performance reported in~\cite{raspireader}. We posit this is for several reasons. First, the results in~\cite{raspireader} are reported on a much smaller dataset (7 materials vs. 12, 15 live subjects vs. 585). Furthermore, the cross-material results in~\cite{raspireader} are reported by leaving only a single material out of training, followed by testing on one withheld material. In contrast the current evaluation protocol only allows training on half of the 12 materials and performs testing on the withheld half. Finally, the results reported in~\cite{raspireader} were at a $FDR$ of 1.0\% which is more lenient than the 0.2\% operating point in the current evaluation.

What is more interesting to note about the cross-material performance of the binary-CNN is that in some cases, the unseen testing spoof materials such as playdoh (Fig.~\ref{fig:playdoh_direct}) or gold finger (Fig.~\ref{fig:gold_direct}) visually look very dissimilar from live fingerprint images (Fig.~\ref{fig:live_direct}). However, since the binary-CNN has not seen the material during training, it frequently misclassifies what intuitively seems like an easy material to classify. In contrast, our proposed GANs can perform very well on materials that look very anomalous such as playdoh and gold fingers (Fig.~\ref{fig:tsne}).

Finally, we acknowledge that the proposed spoof detection system still leaves room for improvement on transparent spoof materials. Indeed, transparent spoofs were also reported as the most challenging materials in~\cite{raspireader} due to the fact that much of the live finger color transmits through the clear spoof materials (Fig.~\ref{fig:failures_gan}).

\vspace{-0.5em}
\section{Summary}
\vspace{-0.25em}
We have introduced a novel one-class classifier built from the discriminators of GAN networks trained on only live fingerprint images for fingerprint spoof detection. We have bested state-of-the-art one-class and binary-class spoof detection algorithms on a large, diverse, and challenging dataset collected with the open source RaspiReader~\cite{raspireader}. Our ongoing research is investigating the fusion of two-class spoof detectors with one-class spoof detectors, followed by further cross-material evaluation.\footnote{This research was supported by the Office of the Director
of National Intelligence (ODNI), Intelligence Advanced
Research Projects Activity (IARPA), via IARPA R\&D Contract
No. 2017 - 17020200004. The views and conclusions
contained herein are those of the authors and should not
be interpreted as necessarily representing the official policies,
either expressed or implied, of ODNI, IARPA, or the
U.S. Government. The U.S. Government is authorized to
reproduce and distribute reprints for governmental purposes
notwithstanding any copyright annotation therein.}

{\small
\bibliographystyle{ieeetr}
\footnotesize{
\bibliography{egbib}}

\begin{thebibliography}{10}

\bibitem{handbook}
D.~Maltoni, D.~Maio, A.~K. Jain, and S.~Prabhakar, {\em Handbook of
  \uppercase{F}ingerprint \uppercase{R}ecognition}.
\newblock Springer, 2nd~ed., 2009.

\bibitem{gummy1}
T.~Matsumoto, H.~Matsumoto, K.~Yamada, and S.~Hoshino, ``Impact of artificial
  "gummy" fingers on fingerprint systems,'' in {\em Optical Security and
  Counterfeit Deterrence Techniques IV}, vol.~4677, pp.~275--290, International
  Society for Optics and Photonics, 2002.

\bibitem{gummy2}
T.~Van~der Putte and J.~Keuning, ``Biometrical fingerprint recognition: don’t
  get your fingers burned,'' in {\em Smart Card Research and Advanced
  Applications}, pp.~289--303, Springer, 2000.

\bibitem{survey}
E.~Marasco and A.~Ross, ``A survey on antispoofing schemes for fingerprint
  recognition systems,'' {\em ACM Computing Surveys (CSUR)}, vol.~47, no.~2,
  p.~28, 2015.

\bibitem{CNN3}
T.~Chugh, K.~Cao, and A.~K. Jain, ``Fingerprint spoof buster: Use of
  minutiae-centered patches,'' {\em IEEE Transactions on Information Forensics
  and Security}, 2018.

\bibitem{inter1}
E.~Marasco and C.~Sansone, ``On the robustness of fingerprint liveness
  detection algorithms against new materials used for spoofing,'' in {\em Int.
  Conf. Bio-Inspired Syst. Signal Process.}, 2011.

\bibitem{inter2}
B.~Tan, A.~Lewicke, D.~Yambay, and S.~Schuckers, ``The effect of environmental
  conditions and novel spoofing methods on fingerprint anti-spoofing
  algorithms,'' {\em {IEEE} {WIFS}}, 2010.

\bibitem{open1}
A.~Rattani, W.~J. Scheirer, and A.~Ross, ``Open set fingerprint spoof detection
  across novel fabrication materials,'' {\em IEEE Trans. on Information
  Forensics and Security}, vol.~10, no.~11, pp.~2447--2460, 2015.

\bibitem{open3}
Y.~Ding and A.~Ross, ``An ensemble of one-class svms for fingerprint spoof
  detection across different fabrication materials,'' in {\em {IEEE}
  International Workshop on Information Forensics and Security}, pp.~1--6,
  2016.

\bibitem{raspireader}
J.~J. Engelsma, K.~Cao, and A.~K. Jain, ``Raspireader: Open source fingerprint
  reader,'' {\em IEEE \uppercase{t}ransactions on \uppercase{p}attern
  \uppercase{a}nalysis and \uppercase{m}achine \uppercase{i}ntelligence}, 2018.

\bibitem{CNN2}
R.~Nogueira, R.~Lotufo, and R.~Machado, ``Fingerprint liveness detection using
  convolutional neural networks,'' {\em {IEEE} Trans. Information Forensics and
  Security}, vol.~11, no.~6, pp.~1206--1213, 2016.

\bibitem{2015}
V.~Mura, L.~Ghiani, G.~L. Marcialis, F.~Roli, D.~A. Yambay, and S.~A.~C.
  Schuckers, ``Livdet 2015-fingerprint liveness detection competition 2015.,''
  in {\em BTAS}, pp.~1--6, 2015.

\bibitem{match}
J.~J. Engelsma, K.~Cao, and A.~K. Jain, ``Fingerprint match in box,'' {\em
  BTAS}, 2018.

\bibitem{dcgan}
A.~Radford, L.~Metz, and S.~Chintala, ``Unsupervised representation learning
  with deep convolutional generative adversarial networks,'' {\em arXiv
  preprint arXiv:1511.06434}, 2015.

\bibitem{mobilenet}
A.~G. Howard, M.~Zhu, B.~Chen, D.~Kalenichenko, W.~Wang, T.~Weyand,
  M.~Andreetto, and H.~Adam, ``Mobilenets: Efficient convolutional neural
  networks for mobile vision applications,'' {\em arXiv preprint
  arXiv:1704.04861}, 2017.

\bibitem{open4}
S.~R. Arashloo, J.~Kittler, and W.~Christmas, ``An anomaly detection approach
  to face spoofing detection: A new formulation and evaluation protocol,'' {\em
  IEEE Access}, vol.~5, pp.~13868--13882, 2017.

\bibitem{open5}
O.~Nikisins, A.~Mohammadi, A.~Anjos, and S.~Marcel, ``On effectiveness of
  anomaly detection approaches against unseen presentation attacks in face
  anti-spoofing,'' in {\em The 11th IAPR International Conference on Biometrics
  (ICB)}, 2018.

\end{thebibliography}
}

\end{document}